\documentclass[11pt,a4paper]{article}
\usepackage{times,latexsym}
\usepackage{url}
\usepackage[T1]{fontenc}
\usepackage{amsfonts}
\usepackage{amssymb}
\usepackage{subcaption,booktabs}
\usepackage{xcolor}
\usepackage[utf8]{inputenc}
\usepackage{hyperref}
\usepackage{inconsolata}
\usepackage{graphicx}
\usepackage{multirow}
\usepackage{stfloats}
\usepackage{appendix}
\usepackage{amsmath}
\usepackage{caption}
\usepackage{tabularray}
\usepackage[ruled, lined, linesnumbered, commentsnumbered, longend]{algorithm2e}
\usepackage[inkscapelatex=false]{svg}

\usepackage[acceptedWithA]{tacl2021v1}
\usepackage{xspace,mfirstuc,tabulary}
\usepackage{arydshln}

\newif\iftaclinstructions
\taclinstructionsfalse 
\iftaclinstructions
\renewcommand{\confidential}{}
\renewcommand{\anonsubtext}{(No author info supplied here, for consistency with
TACL-submission anonymization requirements)}
\newcommand{\instr}
\fi

\iftaclpubformat 

\else

\fi

\title{Cross-Attention is Not Enough: Incongruity-Aware Dynamic Hierarchical Fusion for Multimodal Affect Recognition}

\author{Yaoting Wang$^1$\thanks{\ \ First two authors contributed equally.}\ \ , Yuanchao Li$^1$$^*$, Paul Pu Liang$^2$, \\ \textbf{Louis-Philippe Morency$^2$, Peter Bell$^1$, Catherine Lai$^1$} \\ $^1$University of Edinburgh, $^2$Carnegie Mellon University \\ \texttt{yuanchao.li@ed.ac.uk}}

\date{}

\begin{document}
\maketitle
\begin{abstract}
Fusing multiple modalities has proven effective for multimodal information processing. However, the incongruity between modalities poses a challenge for multimodal fusion, especially in affect recognition. In this study, we first analyze how the salient affective information in one modality can be affected by the other, and demonstrate that inter-modal incongruity exists latently in crossmodal attention. Based on this finding, we propose the Hierarchical Crossmodal Transformer with Dynamic Modality Gating (HCT-DMG), a lightweight incongruity-aware model, which dynamically chooses the primary modality in each training batch and reduces fusion times by leveraging the learned hierarchy in the latent space to alleviate incongruity. The experimental evaluation on five benchmark datasets: CMU-MOSI, CMU-MOSEI, and IEMOCAP (sentiment and emotion), where incongruity implicitly lies in hard samples, as well as UR-FUNNY (humour) and MUStaRD (sarcasm), where incongruity is common, verifies the efficacy of our approach, showing that HCT-DMG: \textbf{1)} outperforms previous multimodal models with a reduced size of approximately 0.8M parameters; \textbf{2)} recognizes hard samples where incongruity makes affect recognition difficult; \textbf{3)} mitigates the incongruity at the latent level in crossmodal attention. (Code available upon acceptance)
\end{abstract}

\section{Introduction}
Emotions are expressed in complex ways  in human communication (e.g., via face, voice, and language). As such, multimodal fusion has become a hot topic in the past decade. Previous studies have shown that by taking advantage of complementary information from multiple modalities, affect recognition can become more robust and accurate \citep{Xu2018,Li2022}. However, several major issues remain unsolved, impeding the progress of Multimodal Information Processing (MIP). First, multimodal signals are not always strictly synchronous. For example, the visual signal usually precedes the audio by around 120ms when people express emotion \citep{grant2001speech}. Second, different modalities may have different or even opposite affective tendencies, which makes affective states difficult to recognize. For instance, people can sometimes say negative content with a positive voice to express politeness \cite{laplante2003things} or smile to express sarcasm \cite{caucci2012social}.

Approaches tackling these issues have been proposed in prior work. For example, \citet{Tsai2019} introduced the Multimodal Transformer (MulT) model to learn a pair-wise latent alignment with the Transformer structure, which directly attends to low-level features in multiple modalities to solve the asynchrony problem. \citet{Wu2021} proposed an incongruity-aware attention network that focuses on the word-level incongruity between modalities by assigning larger weights to words with incongruent modalities. Nevertheless, to capture as much information as possible for better performance, recent models usually repeatedly fuse specific or all modalities \citep{liang2018multimodal}, resulting in not only redundant features but also large model sizes that hinder their real-world use.

To address these problems, in this paper we propose the Hierarchical Crossmodal Transformer with Dynamic Modality Gating (HCT-DMG), a lightweight multimodal fusion model that can alleviate Inter-Modal Incongruity (IMI), reduce information redundancy, and learn representations from unaligned modalities at the same time in a unified framework. Specifically, HCT-DMG dynamically determines the primary modality based on its contribution to the target task and then hierarchically fuses auxiliary modalities via crossmodal Transformers to efficiently obtain the most useful information without modality alignment. The experimental evaluations on CMU-MOSI \citep{Zadeh2016}, CMU-MOSEI \citep{Zadeh2018}, IEMOCAP \citep{Busso2008}, UR-FUNNY \citep{hasan2019ur}, and MUStaRD \citep{castro2019towards} show that our approach achieves highly competitive results with a relatively small model size and alleviates the IMI problem.


\section{Incongruity Analysis}
\label{sec:analysis}
Before presenting our model, we first conduct an incongruity analysis to clarify the IMI problem and why it is necessary to address it, as no previous studies have examined this in detail.

To address the IMI problem, there are typically two approaches: \textbf{1)} Always selecting one modality as the primary input. For instance, sentiment analysis, as well as humor and sarcasm detection have demonstrated the best performance when language is chosen as the primary input \citep{delbrouck2020transformer,rahman2020integrating,ma2023multimodal,hasan2021humor}. However, it cannot be guaranteed that one modality consistently and predominantly contributes to the final task. Even though language is of utmost importance, it does not imply that every sample is dominated by language. \textbf{2)} Employing a weighted sum to adjust the input from each modality, thereby avoiding their equal contributions \citep{rahman2020integrating,yang2020cm}. Yet, this approach overlooks the fact that some modalities can be more closely related or synchronized. Fusing all modalities simultaneously without establishing a hierarchy may diminish the efficacy of fusion. Additionally, the practice of encoding the same input at both the word- and utterance-level for fusion has also been adopted to address incongruity \citep{Wu2021}. However, such an approach may introduce redundancy due to the repeated encoding.

Furthermore, most research typically provides high-level examples to demonstrate how modalities may be misaligned. For instance, in a video clip from MUStaRD for sarcasm detection, a positive sentence, ``You're right, the party is fantastic,'' is presented with a facial expression of ``eye-rolling up'' and drawn-out syllables on the word ``fantastic'' \citep{Wu2021}. However, no evidence has been presented to illustrate how different modalities may mismatch at the latent level. Therefore, we perform an analysis using heatmaps to reveal how crossmodal attention highlights mismatched latent information. We conduct the following experiments on CMU-MOSEI:

\textbf{Exp 1}. Investigates how source modality enhances target modality via crossmodal attention. We use the example of $V\rightarrow T$ (text attended by vision). Next, we hope to see how the combination of two modalities affects the third collectively:

\begin{figure*}[!ht]
    \center{\includegraphics[scale=0.46]{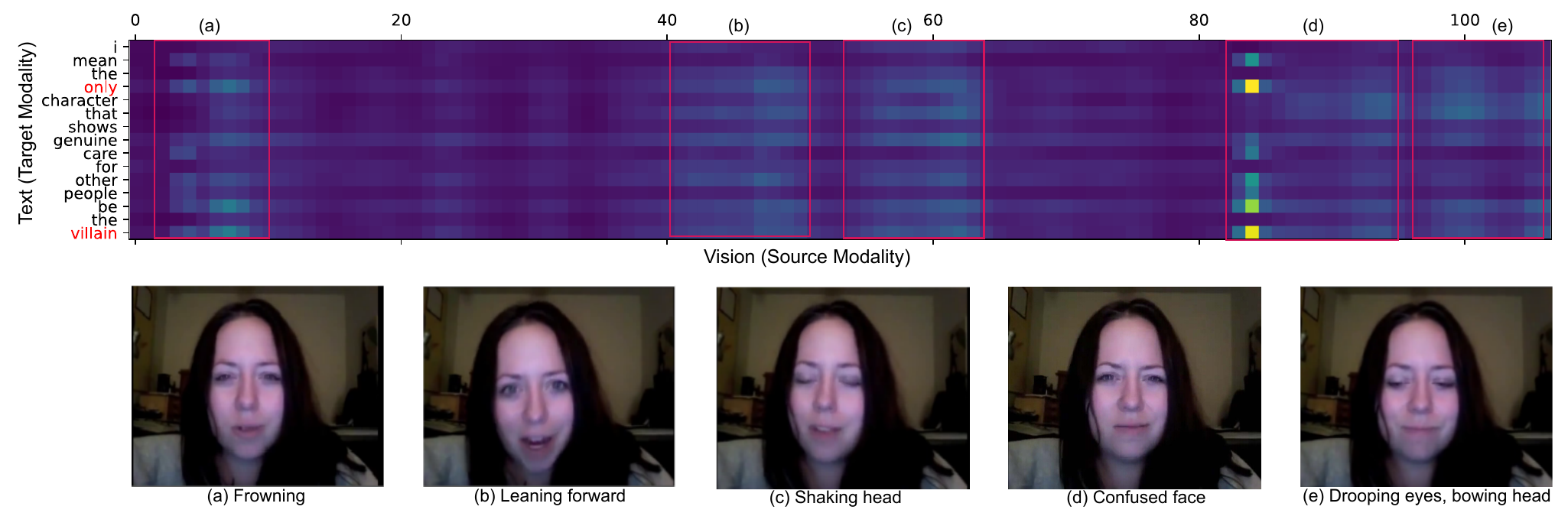}}
    \caption{Exp 1 -- Heatmap of highlighted hidden states using crossmodal attention on $V$ and $T$.}
    \label{exp1}
\end{figure*}

\begin{figure*}[!ht]
    \center{\includegraphics[scale=0.92]{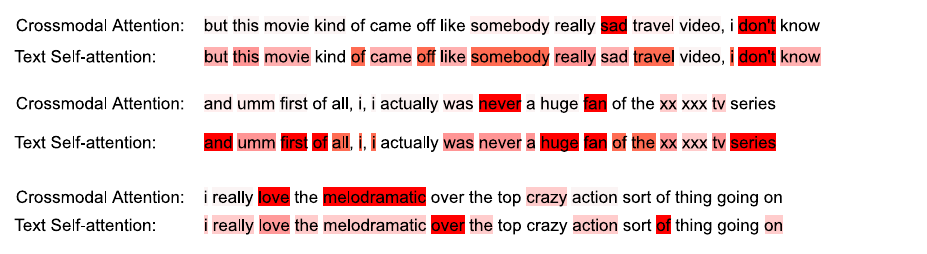}}
    \caption{Exp 2 -- Heatmap of highlighted words using self-attention w/ and w/o crossmodal attention.}
    \label{exp2}
\end{figure*}

\begin{figure*}[!ht]
    \center{\includegraphics[scale=0.44]{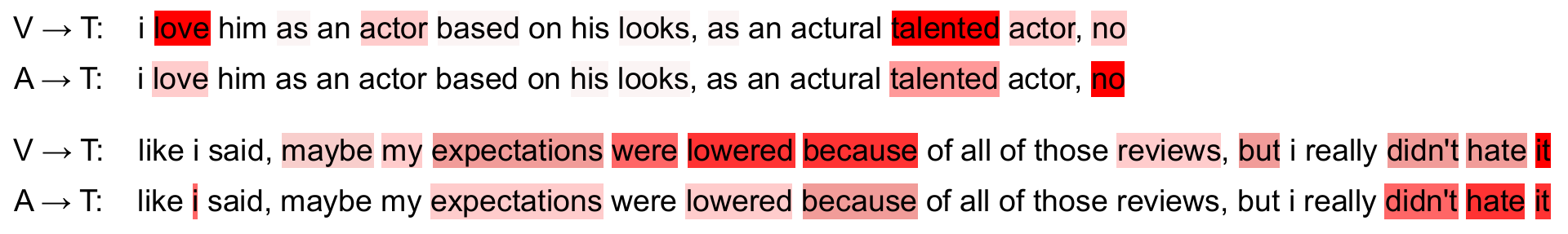}}
    \caption{Exp 3 -- Heatmap of highlighted words using different modalities in crossmodal attention.}
    \label{exp3}
\end{figure*}

\textbf{Exp 2}. Investigates how the salient parts of the target modality are represented by self-attention with and without the combination of source modalities. We use the example of $(A+V)\rightarrow T$ (text attended by crossmodal attention-fused audio-vision). Further, we would like to know how different source modalities affect the target individually:

\textbf{Exp 3}. Investigates how the salient parts of the target modality are represented by crossmodal attention when using different source modalities. We use the examples of $V\rightarrow T$ (text attended by vision) and $A\rightarrow T$ (text attended by audio).

The experimental setup is shown in Table~\ref{setup}, and the visualization is shown in Figure~\ref{exp1}, ~\ref{exp2}, and ~\ref{exp3}.

\begin{table}[ht]
\centering
\caption{Experimental setup for crossmodal attention analysis.}
\scalebox{0.65}{
\begin{tabular}{r|cccc}
\hline
\textbf{Exp.} & \textbf{Target modal} & \textbf{Source modal} & \textbf{Crossmodal} & \textbf{Self-attn}\\ \hline
\larger{1} & \larger{$T$} & \larger{$V$} & \larger{$V\rightarrow T$} & \larger{/} \\
\larger{2} & \larger{$T$} & \larger{$A + V$} & \larger{$(A + V)\rightarrow T$} & \larger{$T$}\\
\larger{3} & \larger{$T$} & \larger{$A$ or $V$} & \larger{$A\rightarrow T$, $V\rightarrow T$} & \larger{/} \\ \hline
\end{tabular}
}
\label{setup}
\end{table}

Figure~\ref{exp1} shows the video frame (x-axis) and text words (y-axis). The salient affective information captured by crossmodal attention is highlighted in red boxes. It can be observed that the highlighted parts are due to obvious facial or behavior changes of the character in the video, such as frowning or shaking head. The crossmodal attention successfully highlights the meaningful words associated with a facial expression (e.g., ``only'', ``villain'').

In Figure~\ref{exp2}, it can be noted that when fused with the combination of source modalities, $T$ focuses more on the words related to emotion with less noise from other words. For example, when with crossmodal attention, the word ``sad'' is the most salient in the first sentence, yet much less focused with self-attention. The same is true for the word ``never'' in the second sentence and the words ``love'' and ``melodramatic'' in the third sentence.

In Figure~\ref{exp3}, we see that when fused with different individual source modalities, the target modality $T$ can be enhanced with disparate affective tendencies. When using $V$ as the source modality, the words ``love'' and ``talented'' are the most highlighted in the first sentence, representing a positive meaning. When using $A$, however, ``no'' is the most focused word, showing negation is important. Similarly, ``but'' showing the turnaround in the second sentence is captured by $V$ yet ignored by $A$, and the two draw attention to different parts. These phenomena demonstrate that different modalities may contain mismatched affective tendencies. The existence of IMI has been found by high-level inter-modal comparison \citep{Desai2022}, sentiment analysis \citep{li2019}, and sarcasm detection \citep{Wu2021}. This analysis indicates that incongruity also exists latently, resulting in hidden states in one modality being affected by the other.

Based on the above findings, we see that crossmodal attention does help multimodal fusion by aligning two modalities to highlight the salient affective information in the target modality with complementary information from the source. According to the attention mechanism \citep{Vaswani2017}, this process can be described as mapping the Query (from the target) to the Key (from the source) and obtain scores for the Value (from the source). However, such a process could malfunction if the modalities have mismatched affective tendencies, which leaves the IMI difficult to resolve at the latent level.

\section{Proposed Approach -- HCT-DMG}
To exploit the advantages of crossmodal attention while solving the above problems, we propose a new multimodal fusion approach: the Hierarchical Crossmodal Transformer with Dynamic Modality Gating (HCT-DMG), which improves on existing methods in two aspects: \textbf{1)} several previous approaches treated all modalities equally and fused them at every step, leaving incongruity in the fusion \citep{Tsai2019,sahay2020low}, while our HCT-DMG approach initially fuses the auxiliary modalities before integrating the primary modality in the final step. This strategy avoids excessive influence on the affect of the primary modality. \textbf{2)} Some prior work determined a primary modality based on the hierarchy of modalities used \citep{rahman2020integrating,hazarika2020misa}. Such a practice is empirical and leads to a fixed hierarchy. So, the weighting pattern (e.g., $T\oplus W_{1}A\oplus W_{2}V$) cannot be changed during model training even though other hierarchies may be better suited to the task. In contrast, HCT-DMG automatically selects and dynamically changes the primary modality in each training batch and constructs the hierarchy accordingly.
Therefore, our proposed approach can eliminate incongruity, reduce redundancy, and enable the model to be modality-agnostic.

\begin{figure*}
    \centering
    \includegraphics[scale=0.54]{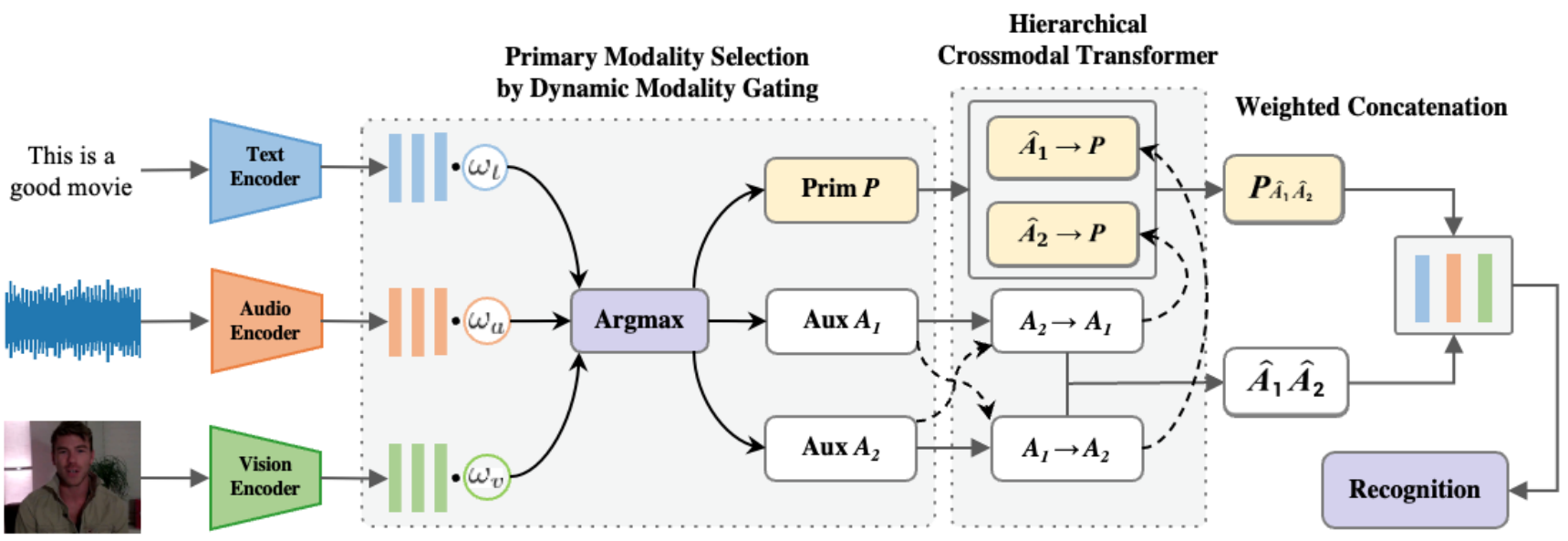}
    \caption{Architecture of HCT-DMG. Prim and Aux are short for primary and auxiliary. Dot lines denote crossmodal attention.}
    \label{HCT-DMG}
\end{figure*}

The architecture is shown as Figure~\ref{HCT-DMG}. HCT-DMG is constructed based on three modalities: Text ($T$), Audio ($A$), and Vision ($V$), and consists of four components: feature encoder, dynamic modality gating, hierarchical crossmodal Transformer, and weighted concatenation. Note that HCT-DMG supports modalities not limited to $T$, $A$, and $V$, as the dynamic modality gating enables to construct the best hierarchy for any three inputs.

\noindent{\textbf{Feature Encoder.}}
\label{sec:feature}
The input features are first fed into 1D Convolutional (Conv1D) networks to integrate local contexts and project the features into the same hidden dimension. Then the features are passed to the Gated Recurrent Unit (GRU) networks, which encode global contexts by updating their hidden states recurrently and model the sequential structure. We use two sets of input features: one uses the same conventional feature extractors as \citet{Tsai2018Learning,Tsai2019} and \citet{sahay2020low} for comparison, while the other uses Large Pre-trained Models (LPM) for performance improvement, which will be described in Section~\ref{sec:eva}.

\noindent{\textbf{Dynamic Modality Gating.}}
DMG determines which modality should be the primary one by the trainable weight for each modality during training, rather than by manual selection. Specifically, each modality is assigned a trainable weight whose value is based on its contribution to the final task. The larger the contribution of a modality, the larger its weight value. The sum of all trainable weights equals to 1, and we allow the weights to be updated in every training batch to ensure that DMG can be well adapted to any type of input modality by working dynamically. We will discuss the details of how DMG works in Section~\ref{sec:mg}.

\begin{algorithm}[!ht]
    \small
    \SetKwFunction{isOddNumber}{isOddNumber}
    \SetKwInOut{KwIn}{Input}

    \KwIn{Primary modality $P$;
    Auxiliary modalities $A_1$ and $A_2$;
    Fixed modality index $Idx_P, Idx_{A_1}, Idx_{A_2} = 0, 1, 2$;
    Dynamic modality gating weight $W$ initialized with \texttt{nn.Param(softmax([1, 1, 1]))}}
   
    \textbf{repeat} \\
    \For{model input}
    {
        
        Sample $Label$, $Text$, $Audio$, $Vision$;
        
        $W' \leftarrow \texttt{softmax(}W\texttt{)}$;

        $P \leftarrow P * W'[Idx_P]$;
        
        $A_1, A_2 \leftarrow A_1 * W'[Idx_{A_1}], A_2 * W'[Idx_{A_2}]$;

        $Idx_{P'} \leftarrow \texttt{argmax(}W'\texttt{)}$;

        \If{$Idx_{P'} == Idx_{A_1}$}
        {
            $P, A_1 \leftarrow A_1, P$
        }
        
        \If{$Idx_{P'} == Idx_{A_2}$}
        {
            $P, A_2 \leftarrow A_2, P$
        }
        
        $Pred \leftarrow \texttt{HCT(}P, A_1, A_2\texttt{)}$;
        $\texttt{update(}Pred, Label\texttt{)}$;
    }
    \textbf{until} no \textit{model input}
    \caption{Working principle of HCT-DMG.}
    \label{alg-dmg}
\end{algorithm}

\noindent{\textbf{Hierarchical Crossmodal Transformer.}}
As a variant of self-attention, cross-attention \citep{lu2019vilbert} transforms the signals from the source modality into a different set of Key-Value pairs to interact with the target modality (in multimodal fusion, cross-attention is usually referred to as crossmodal attention), which has proven useful in various domains \citep{zhang2022cross,rashed2022context}. The crossmodal Transformer used here is the same as MulT \citep{Tsai2019}, which is a deep stacking of several crossmodal attention blocks with layer normalization and positional embeddings. Unlike MulT, which has six crossmodal Transformers in the same step, we use two in the first step to obtain enhanced auxiliary modalities:

\begin{small}
\begin{equation}
    \hat{A}_1=CMT(A_2\rightarrow{A_1})
\end{equation}
\begin{equation}
    \hat{A}_2=CMT(A_1\rightarrow{A_2})
\end{equation}
\end{small}

Then in the second step, another two crossmodal Transformers are used to yield the enhanced primary modality representations:

\begin{small}
\begin{equation}
    \hat{P}_{\hat{A}_1}=CMT(\hat{A}_1\rightarrow{P})
\end{equation}
\begin{equation}
    \hat{P}_{\hat{A}_2}=CMT(\hat{A}_2\rightarrow{P})
\end{equation}
\end{small}

The pseudo code of the working principle of HCT-DMG is shown in Algorithm~\ref{alg-dmg}.

\noindent{\textbf{Weighted Concatenation.}}
After obtaining the enhanced $\hat{P}_{\hat{A}_1}$ and $\hat{P}_{\hat{A}_2}$, we concatenate them and use the self-attention to find its salient parts as the final primary representation:

\begin{small}
\begin{equation}
\hat{P}_{\hat{A}_1\hat{A}_2}=SA(Concat\left[\hat{P}_{\hat{A}_1};\ \hat{P}_{\hat{A}_2}\right])
\end{equation}
\end{small}

At this point, crossmodal representations for every modality have been generated: $\hat{A}_1$, $\hat{A}_2$, and $\hat{P}_{\hat{A}_1\hat{A}_2}$. We concatenate them for the final representation:

\begin{small}
\begin{equation}
Z=Concat\left[W_{1}\hat{A}_1;\ W_{2}\hat{A}_2;\ \hat{P}_{\hat{A}_1\hat{A}_2}\right]
\end{equation}
\end{small}

where $W_{1}$ and $W_{2}$ are the weight matrices, which are learned by the model itself to control how much auxiliary information to extract.

\section{Experiments on Sentiment Analysis and Emotion Recognition}
\label{sec:saer}
We first test our model on sentiment analysis and emotion recognition, as these areas have been extensively researched, yet their incongruity issues have not been well addressed, providing ample baselines for comparison. We describe the datasets and report our results via a comparison with prior models. Since DMG will become obsolete once the primary modality selection converges, we freeze it at that point. We present the process by which HCT-DMG selects the primary modality in Section~\ref{sec:mg}.

\subsection{Datasets and Evaluation Metrics}
\textbf{CMU-MOSI} \citep{Zadeh2016} and \textbf{CMU-MOSEI} \citep{Zadeh2018} are sentiment analysis datasets containing video clips from YouTube, annotated with sentiment scores in the range of [-3, 3]. The former has 2,199 samples, while the latter has 23,454. \textbf{IEMOCAP} \citep{Busso2008} is a multimodal dataset for emotion recognition. Following prior work, we use four emotions (happy, sad, angry, and neutral) for the experimental evaluation, bringing in 4,453 samples.

As with prior work on MOSI and MOSEI, we evaluate the performances using the following metrics: 7-class accuracy (Acc7: sentiment score in the same scale as the labeled scores); binary accuracy (Acc2: positive/negative sentiment polarity); F1 score; Mean Absolute Error (MAE); and the correlation of the recognition results with ground truth (Corr). On IEMOCAP, we report the binary classification accuracy (one versus the others) and F1 score.

\subsection{Experimental Evaluation}
\label{sec:eva}
We use the CMU-SDK \citep{zadeh2018multi}, which splits the datasets into training/validation/testing folds. As described in Section~\ref{sec:feature}, we use both conventional features and LPM features. The conventional features are obtained by using GloVe \citep{pennington2014glove}, FACET\footnote{\url{https://imotions.com/platform/}}, and COVAREP \citep{p56} for $T$, $V$, and $A$, respectively. For the LPM, we use BERT \citep{devlin2018bert} and WavLM \citep{chen2022wavlm} for $T$ and $A$, respectively. We do not use LPM for $V$ as CLIP \citep{radford2021learning}) did not show steady improvement on every dataset as that for $A$ and $T$. The feature and model details are presented in the Appendix.

\subsubsection{Baselines}
\label{sec:5.2.1}
We perform a comparative study against our approach, considering four aspects: \textbf{1)} models using conventional features; \textbf{2)} models using feature from LPM (with $^{\dag}$); \textbf{3)} models using the same crossmodal Transformer as ours (with $^{\diamond}$); \textbf{4)} models with similar sizes to ours (with parentheses showing size). The baselines are as below:

Early Fusion LSTM (\textbf{EF-LSTM}) and Late Fusion LSTM (\textbf{LF-LSTM}) \citep{Tsai2018Learning}. Attention or Transformer-based fusion: \textbf{RAVEN} \citep{Wang2019}, \textbf{MulT} \citep{Tsai2019}. Graph-based fusion: \textbf{Graph-MFN} \citep{Zadeh2018}. Low-rank-based fusion: \textbf{LMF} \citep{Liu2018}. Cyclic translations-based fusion: \textbf{MCTN} \citep{pham2019found}. Context-aware attention-based fusion: \textbf{CIA} \citep{chauhan2019context}. Multi-attention Recurrent-based fusion: \textbf{MARN} \citep{zadeh2018multi}. Temporal memory-based fusion: \textbf{MFN} \citep{zadeh2018memory}. Recurrent multiple stages-based fusion: \textbf{RMFN} \citep{liang2018multimodal}. Low-rank Transformer-based fusion: \textbf{LMF-MulT} \citep{sahay2020low}. Modality-invariant and -specific fusion using LPM: \textbf{MISA} \citep{hazarika2020misa}. We also include several of the above-mentioned models enhanced by Connectionist Temporal Classification (CTC) (cf. \citet{Tsai2019}). 

Note that as the recognition of fine-grained emotions can be significantly improved by word alignment, we do not use the baselines with such alignment for comparison on IEMOCAP. Also, the current State-Of-The-Art (SOTA) results on MOSI and MOSEI are achieved by Self-MM \citep{yu2021learning} and MAG-XLNet \citep{rahman2020integrating}. However, their approaches fundamentally differ from the aforementioned baselines. Self-MM generated additional unimodal labels using multimodal information and labels (i.e., ground-truth) via a self-supervised training scheme. MAG-XLNet directly integrated multimodal information into the Transformer by modifying its structure. Therefore, we exclude them from the comparison since all the models listed in the table concentrate exclusively on fusion methods without introducing extra training tasks or making modifications to the Transformers or pre-trained encoders.

\subsubsection{Results}

\begin{table}[!ht]
\centering
\caption{Comparison results on MOSI, MOSEI, and IEMOCAP. MulT and MISA are our reproduced results using official code. $^{\dag}$: models using feature from LPM. $^{\diamond}$: models using the same crossmodal Transformer as ours.}
\scalebox{0.729}{
\begin{tabular}{l|ccccc}
\hline
\multirow{2}{*}{Model} & \multicolumn{5}{c}{CMU-MOSI} \\
 & Acc7$\uparrow$ & Acc2$\uparrow$ & F1$\uparrow$ & Corr$\uparrow$ & MAE$\downarrow$ \\ \hline
EF-LSTM & 33.7 & 75.3 & 75.2 & 0.608 & 1.023 \\
RAVEN & 33.2 & 78.0 & 76.6 & 0.691 & 0.915 \\
MCTN & 35.6 & 79.3 & 79.1 & 0.676 & 0.909 \\
CTC+EF-LSTM & 31.0 & 73.6 & 74.5 & 0.542 & 1.078 \\
CTC+RAVEN & 31.7 & 72.7 & 73.1 & 0.544 & 1.076 \\
CTC+MCTN & 32.7 & 75.9 & 76.4 & 0.613 & 0.991 \\
MARN & 34.7 & 77.1 & 77.0 & 0.625 & 0.968 \\
MFN & 34.1 & 77.4 & 77.3 & 0.632 & 0.965 \\
RMFN & 38.3 & 78.4 & 78.0 & 0.681 & 0.922 \\
LMF & 32.8 & 76.4 & 75.7 & 0.668 & 0.912 \\
CIA & 38.9 & 79.8 & 79.5 & 0.689 & 0.914 \\
MISA$^{\dag}$ & 41.4 & 81.9 & 81.8 & \textbf{0.762} & \textbf{0.810} \\
MulT$^{\diamond}$ \small{(1.07M)} & 34.3 & 80.3 & 80.4 & 0.645 & 1.008 \\
LMF-MulT$^{\diamond}$ \small{(0.86M)} & 34.0 & 78.5 & 78.5 & 0.681 & 0.957 \\
LF-LSTM \small{(1.24M)} & 33.7 & 77.6 & 77.8 & 0.624 & 0.988 \\ \hdashline
HCT-DMG &  &  &  & \\
\ \ \ \textit{Conven.} \small{(0.78M)} & 39.4 & 82.5 & 82.5 & 0.710 & 0.881 \\
\ \ \ \textit{LPM} \small{(0.83M)} & \textbf{41.8} & \textbf{85.1} & \textbf{84.8} & 0.732 & 0.855 \\ \hline
\end{tabular}
}
\label{mosi}
\end{table}

\begin{table}[!ht]
\centering
\scalebox{0.729}{
\begin{tabular}{l|ccccc}
\hline
\multirow{2}{*}{Model} & \multicolumn{5}{c}{CMU-MOSEI} \\
 & Acc7$\uparrow$ & Acc2$\uparrow$ & F1$\uparrow$ & Corr$\uparrow$ & MAE$\downarrow$ \\ \hline
EF-LSTM & 47.4 & 78.2 & 77.9 & 0.642 & 0.616 \\
RAVEN & 50.0 & 79.1 & 79.5 & 0.662 & 0.614 \\
MCTN & 49.6 & 79.8 & 80.6 & 0.670 & 0.609 \\
CTC+EF-LSTM & 46.3 & 76.1 & 75.9 & 0.585 & 0.680 \\
CTC+RAVEN & 45.5 & 75.4 & 75.7 & 0.599 & 0.664 \\
CTC+MCTN & 48.2 & 79.3 & 79.7 & 0.645 & 0.631 \\
LMF & 48.0 & 82.0 & 82.1 & 0.677 & 0.623 \\
Graph-MFN & 45.0 & 76.9 & 77.0 & 0.540 & 0.710 \\
CIA & 50.1 & 80.4 & 78.2 & 0.590 & 0.680 \\
MISA$^{\dag}$ & 51.8 & \textbf{84.2} & \textbf{84.0} & 0.724 & 0.568 \\
MulT$^{\diamond}$ \small{(1.07M)} & 50.4 & 80.7 & 80.6 & 0.677 & 0.617 \\
LMF-MulT$^{\diamond}$ \small{(0.86M)} & 49.3 & 80.8 & 81.3 & 0.668 & 0.620 \\
LF-LSTM \small{(1.24M)} & 48.8 & 77.5 & 78.2 & 0.656 & 0.624 \\ \hdashline
HCT-DMG &  &  &  &  & \\
\ \ \ \textit{Conven.} \small{(0.78M)} & 50.6 & 81.6 & 81.9 & 0.691 & 0.593 \\
\ \ \ \textit{LPM}  \small{(0.83M)} & \textbf{53.2} & \textbf{84.2} & \textbf{84.0} & \textbf{0.752} & \textbf{0.535}\\ \hline
\end{tabular}
}
\end{table}

\begin{table}[!ht]
\centering
\scalebox{0.611}{
\begin{tabular}{l|cccccccc}
\hline
\multirow{3}{*}{Model} & \multicolumn{8}{c}{IEMOCAP} \\
 & \multicolumn{2}{c}{Happy} & \multicolumn{2}{c}{Sad} & \multicolumn{2}{c}{Angry} & \multicolumn{2}{c}{Neutral} \\
 & Acc & F1  & Acc & F1 & Acc & F1 & Acc & F1 \\ \hline
CTC+EF-LSTM & 76.2 & 75.7 & 70.2 & 70.5 & 72.7 & 67.1 & 58.1 & 57.4 \\
CTC+RAVEN & 77.0 & 76.8 & 67.6 & 65.6 & 65.0 & 64.1 & 62.0 & 59.5 \\
CTC+MCTN & 80.5 & 77.5 & 72.0 & 71.7 & 64.9 & 65.6 & 49.4 & 49.3 \\
MulT$^{\diamond}$ \small{(1.07M)} & 85.6 & 79.0 & 79.4 & 70.3 & 75.8 & 65.4 & 59.5 & 44.7 \\
LMF-MulT$^{\diamond}$ \small{(0.86M)} & 85.6 & 79.0 & 79.4 & 70.3 & 75.8 & 65.4 & 59.2 & 44.0 \\
LF-LSTM \small{(1.24M)} & 72.5 & 71.8 & 72.9 & 70.4 & 68.6 & 67.9 & 59.6 & 56.2 \\ \hdashline
HCT-DMG  &  &  &  & \\
\ \ \ \textit{Conven.} \small{(0.78M)} & 85.6 & 79.0 & 79.4 & 70.3 & 75.8 & 65.4 & 61.0 & 50.5 \\
\ \ \ \textit{LPM} \small{(0.83M)} & \textbf{87.1} & \textbf{81.6} & \textbf{82.4} & \textbf{73.2} & \textbf{79.0} & \textbf{68.8} & \textbf{63.2} & \textbf{60.3} \\ \hline
\end{tabular}
}
\end{table}

The comparison results are shown in Table~\ref{mosi}. On all three datasets, it can be seen that HCT-DMG achieves better results on almost every metric than the baselines when using LPM features, showing the effectiveness of our proposed approach. 

Moreover, when using conventional features, HCT-DMG improves every metric on MOSI and MOSEI, and almost every metric on IEMOCAP compared to the models of similar size. In addition, compared to LMF-MulT and MulT, which use the same crossmodal Transformer as ours, HCT-DMG still outperforms them, especially on MOSI and MOSEI by a large margin. These demonstrate the usefulness of our proposed approach. Furthermore, as HCT-DMG was used to produce the crossmodal attention results in Exp. 2 (in Section~\ref{sec:analysis}), Figure~\ref{exp2} clearly shows that our approach strengthens the most salient affective parts effectively.

Finally, we present some examples in Table~\ref{example}, where incongruity exists. It can be observed that MulT (fusion for six times using the same crossmodal Transformer as ours), fails to handle these difficult cases (raw videos available\footnote{\url{https://sites.google.com/view/taclsubmission}}), producing results that contradict ground truths. In contrast, our approach can recognize true sentiments with very close scores. The examples demonstrate that HCT-DMG can successfully integrate auxiliary modalities with the primary one. The heatmaps of these examples are presented in the Appendix.

\begin{table*}[!ht]
\centering
\caption{Examples containing incongruity from CMU-MOSI.}
\scalebox{0.86}
{
\begin{tabular}{llccc}
\hline
\# & \multicolumn{1}{c}{Spoken words + acoustic and visual behaviors}                                & Ground truth & MulT  & Ours   \\ \hline
1 &
  \begin{tabular}[c]{@{}l@{}}"And that's why I was not excited about the fourth one."\\ + Uninterested tone and facial expression\end{tabular} &
  \colorbox{lime}{-1.4} &
  \colorbox{pink}{1.185} &
  \colorbox{lime}{-1.416} \\ \hline
2 &
  \begin{tabular}[c]{@{}l@{}}"I give Shrek Forever After directed by Mike Mitchell a grade of B minus."\\ + Smile face\end{tabular} &
  \colorbox{pink}{1.0} &
  \colorbox{lime}{-0.576} &
  \colorbox{pink}{0.959} \\ \hline
  3 &
  \begin{tabular}[c]{@{}l@{}}"Um in general um, the little kids seemed to like it that were in there."\\ + Skeptical tone and facial expression\end{tabular} &
  \colorbox{pink}{0.8} &
  \colorbox{lime}{-1.151} &
  \colorbox{pink}{0.700} \\ \hline
4  & \begin{tabular}[c]{@{}l@{}}"I honestly want the aliens to win."\\ + Negative tone (somewhat disdainful) \end{tabular} & 
  \colorbox{lime}{-1.6} & 
  \colorbox{pink}{0.995} & 
  \colorbox{lime}{-1.906}  \\ \hline
  
\end{tabular}
}
\label{example}
\end{table*}

\subsection{Further Analysis and Discussion}
To verify that our approach alleviates the IMI issue as well as to demonstrate how DMG dynamically changes the primary modality, we conduct some further studies (we only present results using conventional features for brevity).

\subsubsection{Resolution of IMI}

\begin{figure}[!ht]
\includegraphics[scale=0.71]{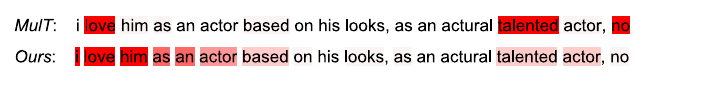}
\caption{A comparison example of IMI.}
\label{incongruity}
\end{figure}

As shown in Figure~\ref{exp3}, the affective words are enhanced by the auxiliary modalities. However, $V$ focuses more on positive words while $A$ highlights negation the most, which likely changes affective tendency. We re-implemented MulT and extracted the attention of its enhanced $T$ (attended by $V$ and $A$) to compare with ours. In Figure~\ref{incongruity}, it can be seen that the positive and negative words are treated equally by MulT, which leaves the IMI unsolved. This is because MulT fuses $V$ and $A$ with $T$ at the same level and simply concatenates two enhanced $T$ modalities. On the other hand, our approach gives little attention to the word ``no'', showing that the IMI is resolved at the latent level as the hidden states of ``no'' are barely encoded. The examples in Table~\ref{example} also demonstrate that the IMI is largely resolved by our approach.

\subsubsection{Automatic Modality Selection by DMG}
\label{sec:mg}
\begin{figure}[!ht]
    \centering
    \includegraphics[scale=0.22]{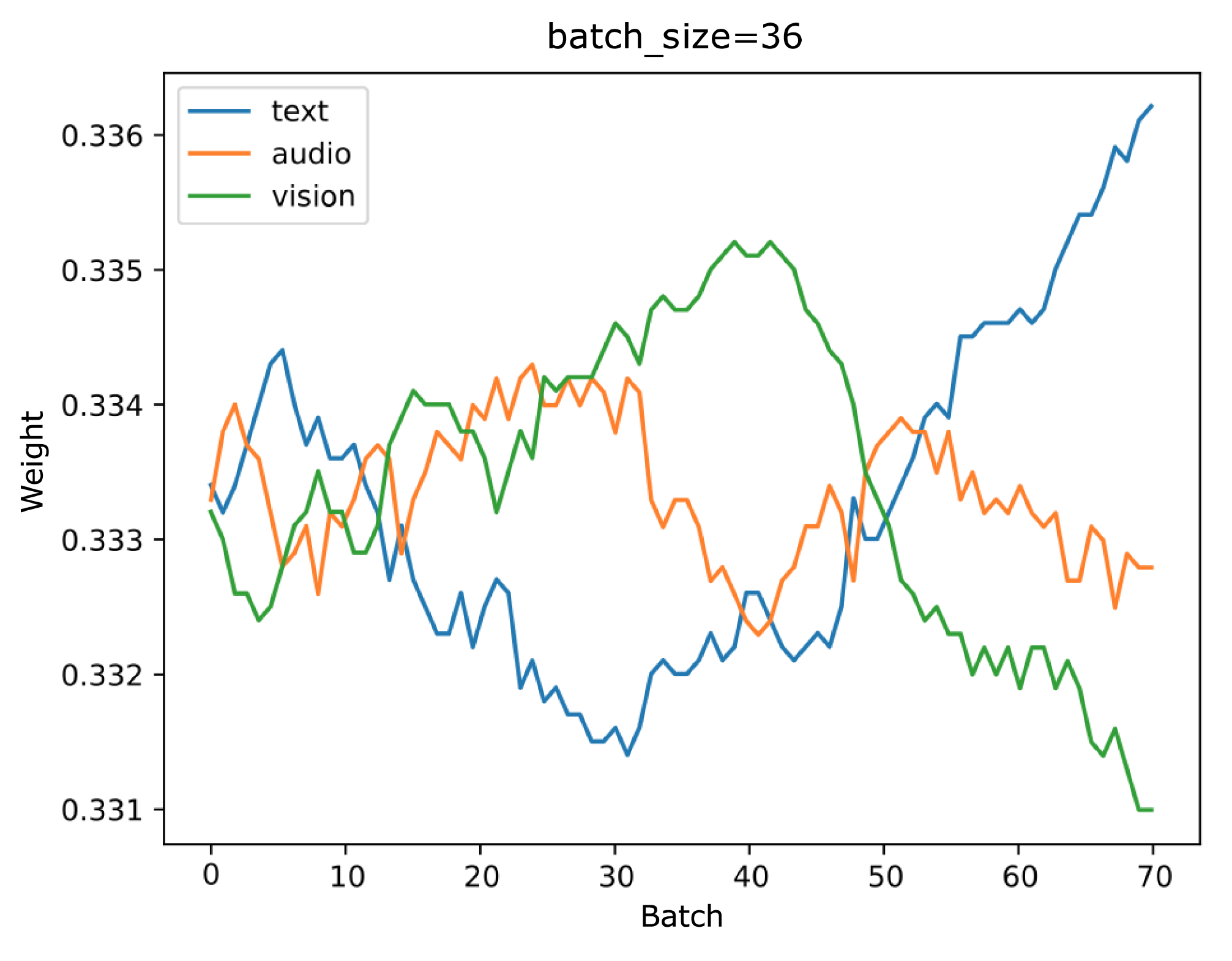}
    \caption{Weight variation in the first epoch on MOSI.}
    \label{fig.16}
\end{figure}

\begin{figure}[!ht]
    \centering
    \includegraphics[scale=0.206]{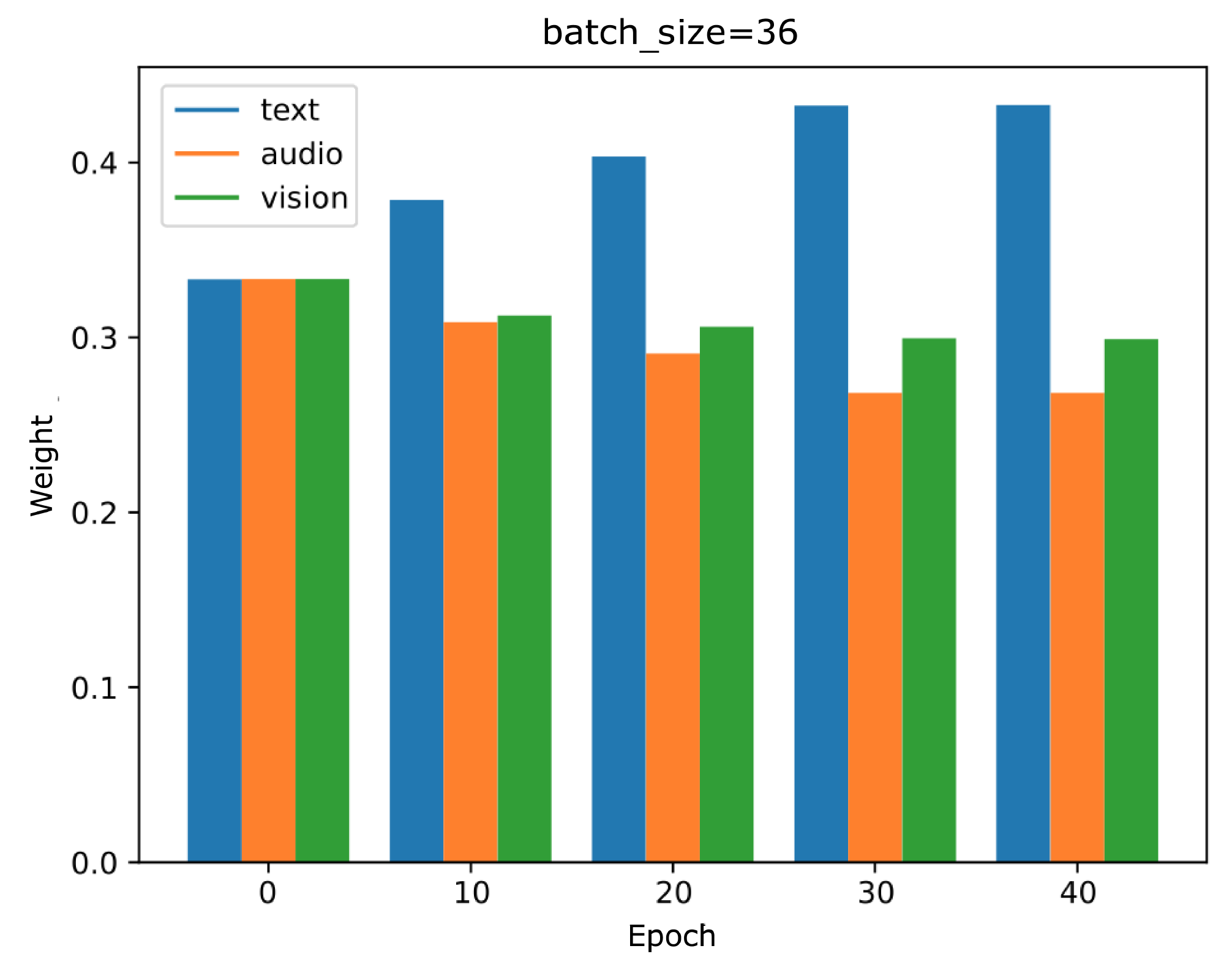}
    \caption{Variation of average weight with epoch on MOSI.}
    \label{fig.17}
\end{figure}

As the DMG automatically selects the primary modality by adjusting the weight for each modality, we show how the weights vary during training using CMU-MOSI. The weight of a modality denotes the confidence that this modality is selected as the primary one. Figure~\ref{fig.16} shows how weights of the modalities vary in the first epoch. It can be noted that $T$ is not the primary one at the beginning but gradually dominates after batch 60. Figure~\ref{fig.17} shows the variation in the average weight of each modality with epoch. It can be seen that $T$ indeed dominates, and the weight distribution starts to converge at around epoch 40. Meanwhile, $V$ gradually surpasses $A$ with epoch although the opposite is true in the first epoch in Figure~\ref{fig.16}.

Note that, the three modalities had very close weights on IEMOCAP  (slightly above 0.33). It is likely due to the fact that IEMOCAP was collected in a lab environment where emotions can be well expressed in all modalities. Furthermore, since HCT-DMG works without the need to understand which modalities are used, its utility can be expected to expand beyond more general scenarios with different signals (e.g., physiological signals). 

\subsubsection{Ablation Study on the Removal of DMG}
\label{sec:ablation}

Although DMG automatically selects $T$ as the primary modality, we would like to see if this phenomenon really brings the best performance compared to prior work whose hierarchies are fixed. Thus, we remove DMG and construct an HCT model with three hierarchies, in each of which either $T$, $A$, or $V$ was manually selected as the primary modality respectively for performance comparison. As shown in Table~\ref{ablation}, selecting $T$ as the primary modality achieves the best performance on every metric on MOSI and MOSEI and on most metrics on IEMOCAP, verifying the rationality and efficacy of the automatic modality selection by DMG.

\begin{table}[!ht]
\centering
\caption{Performance comparison by manually selecting different primary modalities with convention features.}
\label{ablation}
\scalebox{0.835}{
\begin{tabular}{c|ccccc}
\hline
Primary & \multicolumn{5}{c}{CMU-MOSI} \\ 
modality & Acc-7$\uparrow$ & Acc-2$\uparrow$ & F1$\uparrow$ & Corr$\uparrow$ & MAE$\downarrow$ \\ \hline
$T$ & \textbf{38.9} & \textbf{82.5} & \textbf{82.6} & \textbf{0.717} & \textbf{0.859}\\
$A$ & 37.5 & 81.3 & 81.3 & 0.705 & 0.883 \\
$V$ & 38.3 & 80.9 & 81.0 &  0.679 & 0.909 \\ \hline
\end{tabular}
}
\vspace{-5pt}
\label{tab:4}
\end{table}

\begin{table}[!ht]
\centering
\scalebox{0.835}{
\begin{tabular}{c|ccccc}
\hline
Primary & \multicolumn{5}{c}{CMU-MOSEI} \\ 
modality & Acc-7$\uparrow$ & Acc-2$\uparrow$ & F1$\uparrow$ & Corr$\uparrow$ & MAE$\downarrow$ \\ \hline
$T$ & \textbf{50.1} & \textbf{81.8} & \textbf{81.9} & \textbf{0.685} & \textbf{0.601}\\
$A$ & 47.5 & 79.6 & 80.3 & 0.650 & 0.644 \\
$V$ & 48.7 & 80.8 & 81.0 &  0.659 & 0.633 \\ \hline
\end{tabular}
}
\vspace{-5pt}
\end{table}

\begin{table}[!ht]
\centering
\scalebox{0.7}{
\begin{tabular}{c|cccccccc}
\hline
Primary & \multicolumn{8}{c}{IEMOCAP} \\
modality & \multicolumn{2}{c}{Happy} & \multicolumn{2}{c}{Sad} & \multicolumn{2}{c}{Angry} & \multicolumn{2}{c}{Neutral} \\
 & Acc & F1 & Acc & F1 & Acc & F1 & Acc & F1 \\ \hline
$T$ & \textbf{85.6} & \textbf{79.4} & \textbf{79.5} & \textbf{70.6} & 75.8 & 65.4 & 59.6 & \textbf{55.6} \\
$A$ & 85.4 & 79.4 & 78.8 & 70.4 & 75.4 & 65.5 & 59.3 & 52.4 \\
$V$ & 85.6 & 79.2 & 79.4 & 70.3 & \textbf{75.9} & \textbf{65.6} & \textbf{60.4} & 53.3 \\ \hline
\end{tabular}
}
\end{table}

Moreover, although the results of selecting $T$ as the primary modality are the best among the three, overall scores on all three datasets are slightly lower than HCT-DMG's in Table~\ref{mosi}. This phenomenon is reasonable because the primary modality keeps changing during training, even though $T$ is primary overall. Such a dynamic property cannot be encoded by the manual selection of the primary modality whereas it can be encoded by the HCT-DMG and thus yields better results. This is consistent with the weight variation observed in Figure~\ref{fig.16} and \ref{fig.17}, which reaffirms the validity of our approach. In addition, the results on IEMOCAP, where no modality consistently dominates, align with their close weights (slightly above 0.33) as discussed in Section~\ref{sec:mg}.

\section{Experiments on Humor and Sarcasm Detection}
To elaborate on our findings, we conduct additional experiments on humor and sarcasm detection. Unlike sentiment analysis and emotion recognition, where incongruity is implicit, humor and sarcasm expressions commonly contain explicit incongruity across modalities. Therefore, these two tasks serve as ideal benchmarks for assessing the effectiveness and generalizability of our proposed approach.

\subsection{Datasets and Evaluation Metrics}
\textbf{UR-FUNNY} \cite{hasan2019ur} is collected from TED talk videos, incorporating language, acoustic, and visual modalities, as well as the context preceding the punchline. The punchline is extracted using the `laughter' markup, indicating when the audience laughed during the talk, in the transcripts. The sentences preceding the punchline form the context. Negative samples are also extracted in a similar manner, where target punchline utterances are not followed by `laughter'. In total, the dataset consists of 5,000 humor and 5,000 non-humor instances from 1,741 distinct speakers. We use its version 2, following the baseline works.

\textbf{MUStARD} \citep{castro2019towards} is sourced from popular TV shows such as Friends, Big Bang Theory, Golden Girls, and Sarcasmaholics. This dataset comprises 690 video segments that have been manually annotated with labels indicating sarcasm or non-sarcasm. For each segment, the dataset includes the target punchline utterance and the associated historical dialogues as context.

We report the binary classification accuracy (Acc2: humor/non-humor or sarcastic/non-sarcastic) following the literature.

\subsection{Experimental Evaluation}
The training, validation, and testing sets have already been split and provided by the datasets. Same as the experiments in Section~\ref{sec:saer}, we use both conventional and LPM features. The feature and model details are presented in the Appendix.

\subsection{Baselines}
We compare HCT-DMG with the following baselines: Memory fusion by incorporating the information from the preceding context: \textbf{C-MFN} \citep{hasan2019ur}. Modality-invariant and -specific fusion: \textbf{MISA} \citep{hazarika2020misa} (we also include their variants that used different language models: Glove, BERT, ALBERT \cite{lan2019albert}). Integration of the preceding context and external knowledge: \textbf{HKT} \citep{hasan2021humor} (we use the performance without its proposed humor centric features because these features represent an additional modality, resulting in an unfair comparison with the others). As the same reason in Section~\ref{sec:5.2.1}, we do not include MAG-XLNet for comparison. Since humor and sarcasm detection are relatively new tasks in affective computing compared to sentiment analysis and emotion recognition, there are fewer baselines available. Therefore, we also include some of their ablation models. Since there are no small models, we omit the size comparison.

\subsection{Results and Discussion}

The comparison results are shown in Table~\ref{urfunny}. On both datasets, it can be observed that HCT-DMG achieves superior results compared to most of the baselines when using LPM features, demonstrating the effectiveness of our proposed approach. Furthermore, even when using conventional features, which are not very powerful, we also obtain highly competitive performance compared to some baseline models using LPM.

\begin{table}[!ht]
\centering
\caption{Comparison results on UR-FUNNY and MUStARD for humor and sarcasm detection. \\ $^{\dag}$: models using feature from LPM.}
\scalebox{0.75}{
\begin{tabular}{l|cc}
\hline
\multirow{2}{*}{Model} & \multicolumn{1}{c}{UR-FUNNY} & \multicolumn{1}{c}{MUStARD} \\
 & Acc2$\uparrow$ & Acc2$\uparrow$ \\ \hline
C-MFN (GloVe) & 65.23 & - \\
C-MFN (ALBERT)$^{\dag}$ & 61.72 & - \\
MISA (GloVe) & 68.60 & - \\
MISA (BERT)$^{\dag}$ & 69.62 & 66.18 \\
MISA (BERT)$^{\dag}$ [punchline only] & 70.61 & - \\
MISA (ALBERT)$^{\dag}$ & 69.82 & 66.18 \\
HKT$^{\dag}$ & \textbf{76.36} & 75.00 \\
HKT$^{\dag}$ ($T$ only) & 73.54 & 73.53 \\
HKT ($A$ only) & 64.99 & 73.53 \\
HKT ($V$ only) & 55.84 & 64.71 \\

\hdashline
HCT-DMG &  &   \\
\ \ \ \textit{Conventional} & 68.54 & 73.29 \\
\ \ \ \textit{LPM} & 75.09 & \textbf{76.62} \\ \hline
\end{tabular}
}
\label{urfunny}
\end{table}

To the best of our knowledge, HKT is the SOTA model on these two tasks in the literature thus far. Our result on UR-FUNNY does not surpass theirs, but we achieve higher result on MUStARD, demonstrating the competitiveness of our model compared to the SOTA. Note that HKT was developed specifically for experimentation on the two datasets without being tested on MOSI, MOSEI, and IEMOCAP, there is no evidence that it can perform as well as on sentiment analysis and emotion recognition. On the contrary, HCT-DMG performs well on all five datasets, demonstrating that our proposed approach addresses IMI not only in sentiment and emotion but also in more specific scenarios such as humor and sarcasm. Therefore, the generalizability of HCT-DMG is well-established, allowing for its extension to various MIP tasks.

\subsection{Automatic Modality Selection by DMG}

\begin{figure}[!ht]
    \centering
    \includegraphics[scale=0.21]{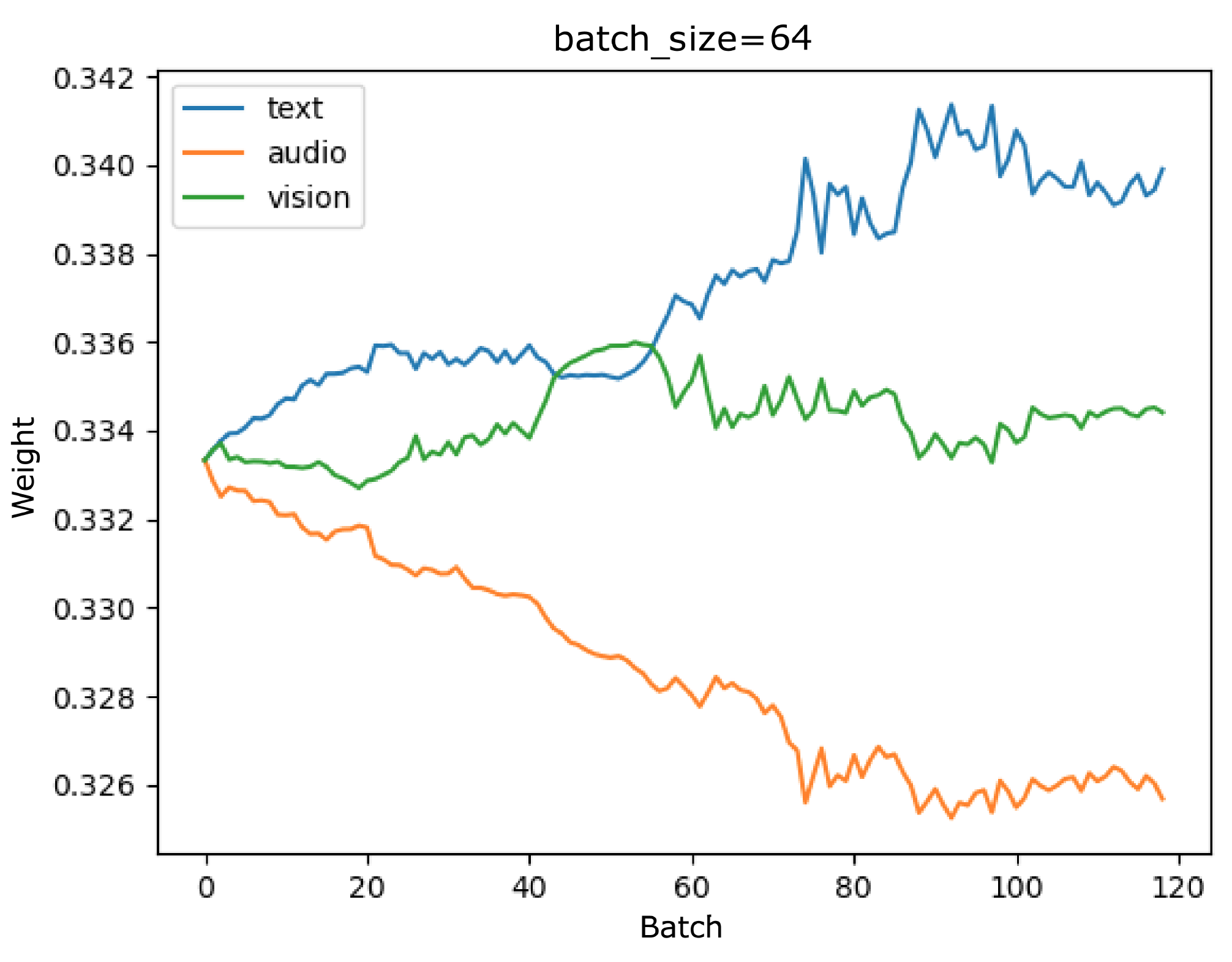}
    \caption{Weight variation in the first epoch on UR-FUNNY.}
    \label{fig.urf_bs64}
\end{figure}

\begin{figure}[!ht]
    \centering
    \includegraphics[scale=0.207]{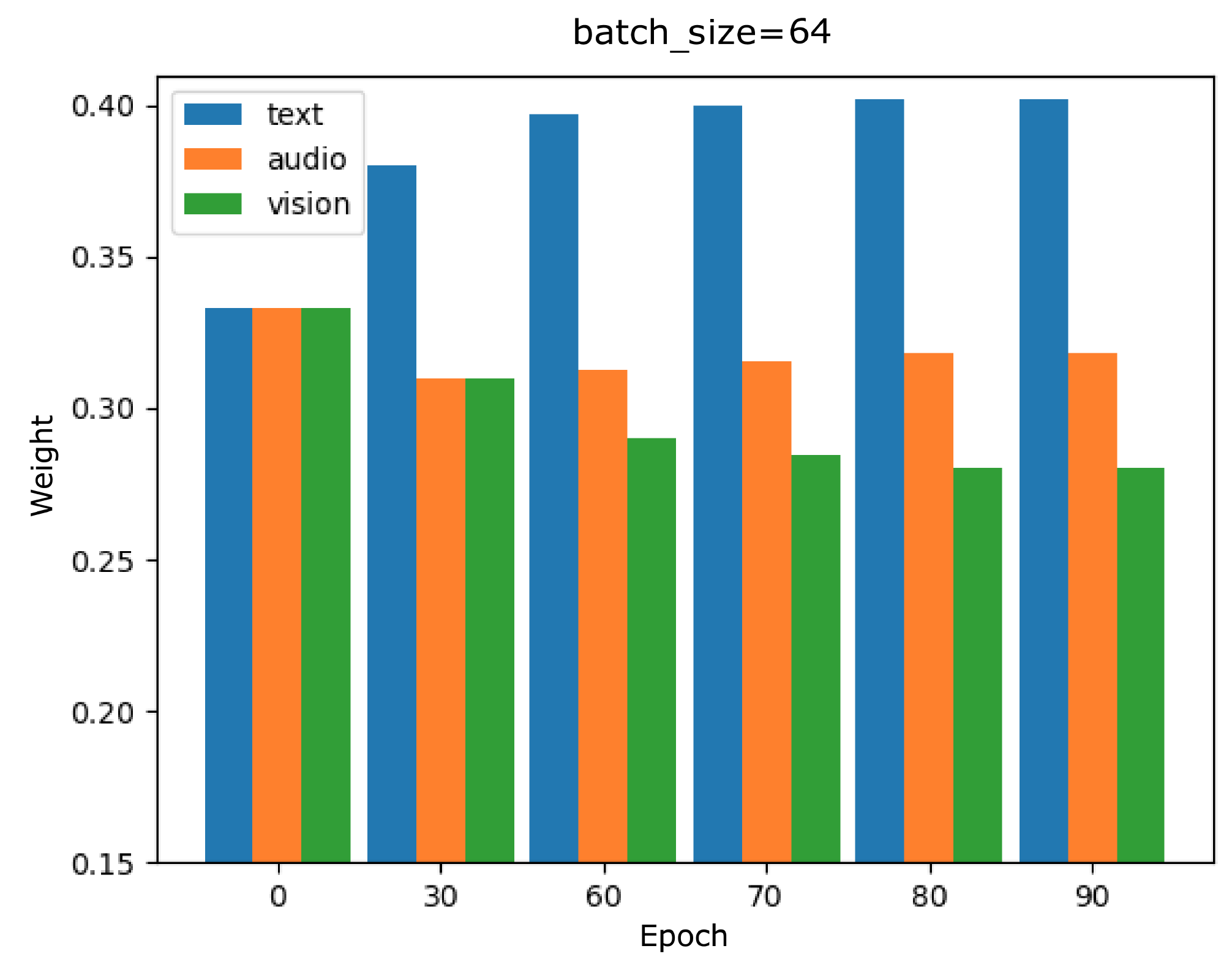}
    \caption{Variation of average weight with epoch on UR-FUNNY.}
    \label{fig.urf_prob}
\end{figure}

Unlike the weight variation on MOSI, the process of automatic modality selection by DMG exhibits different patterns on UR-FUNNY. Figure~\ref{fig.urf_bs64} shows that $T$ is selected as the primary modality from the very beginning, and $V$ appears to be more crucial than $A$. With the training epoch, the weight of $A$ starts to surpass that of $V$ as shown in Figure~\ref{fig.urf_prob}. While on MOSI, the opposite is true: Figure~\ref{fig.16} shows that the modalities compete with each other for about 50 batches until $T$ steadily dominates, and $A$ is seen as more crucial than $V$, and Figure~\ref{fig.17} tells that with the training goes, the weight of $V$ starts to surpass that of $A$.

From these interesting findings, we can infer that \textbf{1)} $T$ plays a crucial role in both sentiment analysis and humor detection. \textbf{2)} $V$ conveys more sentiment information than $A$ whereas $A$ accounts more for humor. \textbf{3)} Even though the weights seem to be steady in the first epoch ($V$ > $A$), DMG can still dynamically change them as the training goes on  ($A$ > $V$), demonstrating the efficacy of our approach.

Here, we omit the ablation study of removing DMG as Figure~\ref{fig.urf_prob} is sufficient to show their individual contribution supported by the consistency between Table~\ref{tab:4} and Figure~\ref{fig.17} on MOSI. It is highly likely that the manual selection will present the contribution in the order of $T$ > $A$ > $V$ here. The prior work by HKT also explored the contribution of each modality. This was accomplished through ablation studies, where each modality was individually used, and the other two were removed \citep{hasan2021humor}. They observed that in humor detection, the contribution of each modality followed the order of $T$ > $A$ > $V$, which is the same as our finding, further suggesting that the principle of our DMG is reasonable. In the context of sarcasm detection, they noticed that $T$ and $A$ contributed almost equally, with $V$ having the least impact. This is plausible since humor primarily originates from textual information, particularly punchlines, while sarcasm is fundamentally a combination of language and tone, with neither heavily relying on facial expressions.

\subsection{Effect of Batch Size}
Since our model dynamically selects the primary modality for each training batch (i.e., all samples in the same batch share the same primary modality), the choice of batch size will have an impact on the training process and performance. Therefore, we explore various batch sizes to determine their effect on achieving the optimal performance. The comparison results are shown in Table~\ref{batch}.

\begin{table}[!ht]
\centering
\caption{Comparison results on UR-FUNNY and MUStARD using different batch sizes.}
\scalebox{0.9}{
\begin{tabular}{c|cc}
\hline
\multirow{2}{*}{Batch size} & \multicolumn{1}{c}{UR-FUNNY} & \multicolumn{1}{c}{MUStARD} \\
 & Acc2$\uparrow$ & Acc2$\uparrow$ \\ \hline
1 & 68.70 & 72.71 \\
4 & 71.95 & 74.40 \\
16 & 73.38 & 74.97 \\
32 & 74.03 & 75.02 \\
64 & \textbf{75.09} & 75.31 \\
128 & 74.26 & 74.88 \\
256 & 74.03 & \textbf{76.62} \\ 
512 & 74.65 & 75.31 \\
\hline
\end{tabular}}
\label{batch}
\end{table}

We surprisingly observe that batch size largely impacts performance, with 64 being the best for humor detection and 256 for sarcasm. The performance difference across different batch sizes can be substantial. Usually, batch size does not make such a significant difference \cite{smith2017don,lian2022smin}, demonstrating that our model indeed learns from the dynamic modality selection over each batch. We also present a visualization comparison of how DMG selects the primary modality with different batch size in the Appendix.

We omit the batch size effect for sentiment and emotion recognition, as we observed that using a large batch size hindered their accuracy. For MOSI, MOSEI, and IEMOCAP, we found that the optimal batch sizes are 32, 64, and 16, respectively. This is plausible because multi-class problems, particularly in the context of categorical emotion recognition, should benefit from smaller batch sizes to ensure precise primary modality selection.

\section{Discussion on Primary Modality in Multimodal Affect Recognition}
At last, we have observed that, regardless of the variations in the contributions of $V$ and $A$ across different datasets, $T$ consistently maintains its prominence overall. To the best of our knowledge, there is no detailed explanation as to why choosing $T$ as the primary modality works the best for MIP, especially for affect recognition tasks. Hence, we gather the following empirical findings from different perspectives, including linguistics, neuroscience, speech production, and language perception, for a comprehensive discussion.

\textbf{1)} There is a clear temporal pattern when people express emotions via vision and audio modalities: visual signals usually precede audio by around 120ms \citep{grant2001speech}. Thus, fusing these two modalities usually produces stable enhanced information to $T$ \citep{Li2020}.

\textbf{2)} People can behave quite differently from what they say in spoken dialogues. For example, positive behaviors sometimes come along with a negative sentence to ease the embarrassment \citep{li2019}, and a positive sentence can be said in a negative voice to express sarcasm \citep{castro2019towards}. Hence, $T$ is usually treated as the target modality to manipulate with $A$ and $V$.

\textbf{3)} In emotion recognition applications, a misrecognition of emotion at the level of sentiment polarity would lead to a fatal error (imagine that the system responds ``Good to hear that!'' in a happy voice when the user in fact feels sad). On the other hand, misclassifying an emotion as another that has the same polarity may well be tolerable \citep{tokuhisa2008emotion}. Given the fact that sentiment largely depends on $T$ \citep{Lindquist2015}, its role is highlighted as $T$ anchors the most significant affective tendency.

\textbf{4)} Using $T$ as the primary modality in fine-tuning and shifting the language-only position of a word to the new position in light of $AV$ information allows the language models (e.g., BERT, XLNet) to better yield sentiment scores \citep{rahman2020integrating}. In conjunction with the conclusion in point 3), this demonstrates that employing $AV$ to assist $T$ is an optimal choice.

\textbf{5)} Modality refers to the way in which something is expressed or perceived. Unlike $A$ and $V$, which are raw (low-level) modalities closest to sensors, $T$ is a relatively abstract and high-level modality that is farther from sensors \citep{baltruvsaitis2018multimodal}. According to the nature of the human brain's hierarchical perceptual processing, low-level information is processed first, followed by high-level information \citep{peelle2010hierarchical}. Thus, using a hierarchical model to process low-level features and fuse high-level ones sequentially can yield better representations for MIP, especially multimodal affect recognition \citep{tian2016recognizing,Li2022}.

We expect that our proposed HCT-DMG will bring new insights to the literature and, together with the aforementioned studies, provide a theoretical basis to support the conclusion that \textit{text is relatively independent from audio and vision yet significantly contributes to affective states.}

\section{Related Work}
Models utilizing data from different modalities usually outperform unimodal ones as more information is aggregated. Prior work has shown that learning with multiple modalities is superior to employing a subset of modalities, since the former has access to a better latent space representation \citep{huang2021makes}. Among previous approaches, early fusion and late fusion are the most widely used for MIP. However, due to the strict constraint on time synchrony, early fusion does not work well if the input features of multiple modalities differ in their temporal characteristics \citep{Li2020}. On the other hand, since different modalities have been confirmed to be complementary to each other \citep{chuang2004multi}, the relatedness among them is ignored by late fusion. To this end, tensor fusion, which is performed at the latent level, has become mainstream. For example, \citet{zadeh2017} introduced a Tensor Fusion Network, that learns both intra- and inter-modality dynamics end-to-end. 

With the success of the cross-attention mechanism \citep{lu2019vilbert}, which exchanges key-value pairs in self-attention, a major trend using cross-attention for multimodal fusion has emerged. \citet{Tsai2019} proposed a crossmodal attention-based Transformer to provide tensor-level crossmodal adaptation that fuses multimodal information by directly attending to features in other modalities. \citet{Zadeh2019} developed a self-attention- and cross-attention-based Transformer to extract intra-modal and inter-modal emotional information, respectively. \citet{Li2022} used crossmodal attention with a hierarchical structure to capture lexical features from different textual aspects for speech emotion recognition.

However, there is no guarantee that using multimodal data is always better than unimodal. For example, \citet{huang2021makes} found that combining multiple modalities (text, audio, and video) underperforms the unimodal when sample sizes are relatively small. Moreover, \citet{rajan2022cross} compared a self-attention and a cross-attention model for emotion recognition, showing no clear difference between the results of the two models.

One of the possible reasons behind these phenomena is the incongruity, i.e., the mismatched affective tendencies across different modalities, resulting in IMI -- a general problem for MIP tasks. Incongruity has been recognized as one of the primary mechanisms for generating humor and sarcasm \citep{stock2003getting}, as well as implicit sentiment and emotion such as cold anger \citep{yacoub2003recognition}. 

The majority of the previous research on IMI was based on high-level comparison analysis between modalities, such as a person expressing praise while rolling his/her eyes \citep{Wu2021}. However, as discussed previously, little work had been done on how IMI could be effectively addressed at the latent level. 
Additionally, it is a common practice in multimodal fusion research to repeatedly fuse certain or even all modalities to exploit as much information as possible. For instance, the widely recognized MulT model \citep{Tsai2019} performed crossmodal fusion six times on three modalities, with each modality fused with the other two separately. Such an operation would bring information redundancy to the model and result in large model sizes, which hinder the real-world use of MIP.

With these challenges in mind, we presented HCT-DMG, a lightweight incongruity-aware model, which dynamically chooses the primary modality in each training batch and reduces fusion times by leveraging the learned hierarchy in the latent space.

\section{Conclusions}
In this study, we analyze crossmodal attention-based multimodal fusion and propose a hierarchical crossmodal Transformer with dynamic modality gating for incongruity-aware multimodal affect recognition. The major contributions are:

\textbf{1)} We demonstrate the existence of inter-modal incongruity at the latent level due to crossmodal attention. Specifically, we show that crossmodal attention can help to capture affective information across modalities and enhance salient parts in the target modality, but it can also induce mismatched affective tendencies from different modalities.

\textbf{2)} We propose a hierarchical crossmodal Transformer with dynamic modality gating -- HCT-DMG, which automatically selects the primary modality during training. This model requires fewer fusion operations and does not repeatedly fuse a single modality, reducing the parameters to approximately 0.8M while significantly outperforming existing models of similar size.

\textbf{3)} We further analyze the mechanism and feasibility of automatic modality selection by DMG and show that the selection process supports the primacy of text in prior multimodal information processing studies, adding new insights to the literature.

\textbf{4)} We test the performance of HCT-DMG on five datasets for sentiment analysis, emotion recognition, humor and sarcasm detection. HCT-DMG achieves remarkable results, consistently outperforming almost all of the baseline models across all tasks. It highlights the versatility and effectiveness of our proposed approach in handling various multimodal affect recognition tasks. 

In our future work, we plan to evaluate HCT-DMG in other multimodal domains to create a comprehensive benchmark. We will also explore scenarios in which other modalities, such as physiological signals, are present, or errors exist in $T$, such as using transcribed results from speech recognition. Besides, it would interesting to investigate if there are differences in the primary modality when experimenting on datasets, such as RAVDESS \citep{livingstone2018ryerson}, where $T$ has been proven to barely contribute to affect recognition. Additionally, we aim to establish a general hierarchy for handling more than three modalities. Finally, we intend to investigate dimensionality reduction techniques \citep{shao2023erasure} to further reduce unaligned information that may result in redundant and misleading information from the learned representations.

\section*{Limitations}
As with all the other supervised learning tasks, our model relies on the accuracy of the labels. Nevertheless, it is challenging to label difficult cases in situations where inter-modal incongruity, ambiguous emotions, or missing information are present. Take the cases in Table~\ref{example} as an example: 1) In our opinion, \#2 should be a neutral emotion with the value of 0, as the person is just stating a fact. 2) There is a word ``but'' missing at the end of the labeled sentence of \#3 (can be clearly noticed in the audio or the video), which is a sign indicating a turnaround in attitude. It can be regarded as the same situation as the sample in Figure~\ref{incongruity}. However, \#3 separates the whole sentence into two sub-sentences, while the sample in Figure~\ref{incongruity} combines two sub-sentences as a whole. Such an inconsistency in labeling introduces incongruity and ambiguity into the tasks and hinders the training of robust and applicable models.

Besides, although the inter-modal incongruity is largely removed by the hierarchical architecture, the affective tendency could be wrong if ambiguity exists in the primary modality. In Figure~\ref{incongruity}, our approach recognizes this sample as positive with the score of \textit{+1.84}, but the ground truth is labeled as negative with the score of \textit{-1}. It is likely because the first half of the text denotes a positive sentiment, yet the second is obviously negative. Without contextual knowledge, it is almost impossible for a system to know that the second half is the focus of the content, as humans can.

\bibliography{tacl2021}
\bibliographystyle{acl_natbib}

\appendix

\section{Appendix}
\label{sec:appendix}

\subsection{Datasets}
\begin{table}[!ht]
\caption{Data distribution and modality sampling rate of CMU-MOSI and CMU-MOSEI. {$S_{A}$} for audio sampling rate and {$S_{V}$} for vision sampling rate.}
\centering
\scalebox{0.74}{
\begin{tabular}{l|cccccc}
\hline
\textbf{Dataset} & \textbf{Train} & \textbf{Valid} & \textbf{Test} & \textbf{Total} & \textbf{$S_{A}$} & \textbf{$S_{V}$} \\ \hline
CMU-MOSI & 1284 & 229 & 686 & 2199 & 12.5 & 15\\
CMU-MOSEI & 16,326 & 1871 & 4659 & 22,856 & 20 & 15\\ \hline
\end{tabular}
}
\label{table.1}
\end{table}

\begin{table}[!ht]
\centering
\caption{Data distribution of four emotions in the IEMOCAP dataset.}
\scalebox{0.78}{
\begin{tabular}{l|cccc}
\hline
\textbf{Emotions} & \textbf{Train} & \textbf{Valid} & \textbf{Test} & \textbf{Total} \\ \hline
Neural & 954 & 358 & 383 & 1695 \\
Happy & 338 & 116 & 135 & 589 \\
Sad & 690 & 188 & 193 & 1071 \\
Angry & 735 & 136 & 227 & 1098 \\
Total & 2717 & 798 & 938 & 4453 \\ \hline
\end{tabular}
}
\label{table.2}
\end{table}

\subsection{Extracted Features}

The sequence lengths and feature dimensions of the three modalities in the three benchmarks are shown in Table \ref{table.3}.

\begin{table}[!ht]
\centering
\caption{Sequence lengths and feature dimensions of the three modalities in the three benchmark datasets. $dim$: Conventional features. $dim_l$: Features extracted from LPM. *: The development team screened the vision and audio features of CMU-MOSI.}
\scalebox{0.68}{
\begin{tabular}{l|cccccccc}
\hline
\textbf{Dataset} & \multicolumn{3}{c}{\textbf{Text}} & \multicolumn{2}{c}{\textbf{Vision}} & \multicolumn{3}{c}{\textbf{Audio}} \\
 & len & dim & dim$_l$ & len & dim & len & dim & dim$_l$ \\ \hline
CMU-MOSI & 50 & 300 & 768 & 500 & 20* & 375 & 5* & 768 \\
CMU-MOSEI & 50 & 300 & 768 & 500 & 35 & 500 & 74 & 768 \\
IEMOCAP & 20 & 300 & 768 & 500 & 35 & 400 & 74 & 768 \\ 
UR-FUNNY & 130 & 300 & 768 & 130 & 371 & 130 & 81 & 768 \\ 
MUStARD & 12 & 300 & 768 & 200 & 2048 & 30 & 283 & 768 \\ 
\hline
\end{tabular}
}
\label{table.3}
\end{table}

\noindent{\textbf{Textual Features: a) GloVe.}} In convention, the transcriptions in all three datasets use the global word embeddings generated by GloVe in convention. This distributed representation allows words in the same context to be close to each other in the vector space and maintain specific relationships. For this pre-extracted data, the text modal features are trained and derived from 840 billion tokens with 300 dimensions of GloVe embeddings. \noindent{\textbf{b) BERT.}} BERT is a pre-trained language model based on the Transformer architecture and trained on massive amounts of text data using unsupervised learning. BERT has achieved highly desirable results on a wide range of natural language processing tasks, including question answering, sentiment analysis, and natural language inference. \noindent{\textbf{c) ALBERT.}} ALBERT (A Lite BERT) is a pre-trained language model that builds upon the BERT model with the goal of reducing model size and computational requirements while maintaining high performance. ALBERT is designed for efficiency and scalability, making it capable of running in resource-constrained environments and easily scaling to larger datasets and tasks.

\noindent{\textbf{Vision Features: FACET.}} FACET is a commercial facial emotion detection software developed by iMotions. The software can demonstrate 35 facial action units and record facial muscle movements to represent frame-by-frame emotions.

\noindent{\textbf{Audio Features: a) COVAREP.}} COVAREP is an open-source repository for speech processing, supporting collaboration and free access. The features of the processed speech data are based on pitch tracking, polarity detection, spectral envelopes, glottal flow, and other common speech features \citep{p56}. The pre-extracted data contains 74 dimensions of speech features. \noindent{\textbf{ b) WavLM.}} WavLM is a pre-trained speech model that simultaneously learns masked speech prediction and denoising during pre-training, resulting in significant improvements for a range of speech processing tasks. 

\subsection{Hyperparameters Tuning}
After tuning the hyperparameters, we find the optimal settings, as shown in Table \ref{table.6}.

\begin{table}[!ht]
\centering
\caption{Hyperparameter settings for the five datasets.}
\scalebox{0.56}{
\begin{tabular}{l|ccccc}
\hline
\textbf{Setting} & \textbf{MOSI} & \textbf{MOSEI} & \textbf{IEMOCAP} & \textbf{UR-FUNNY} & \textbf{MUStARD} \\ \hline
learning rate & 1e-3 & 1e-3 & 1e-5 & 1e-3 & 1e-3 \\
batch size & 36 & 64 & 16 & 64 & 256 \\
hidden size & 40 & 40 & 40 & 40 & 40 \\
kernel (T/A/V) & 1/1/1 & 1/1/1 & 1/1/1 & 1/1/1 & 1/1/1 \\
number of epochs & 30 & 30 & 60 & 60 & 60 \\
transformer layers & 4 & 4 & 4 & 4 & 4 \\
attention heads & 5 & 5 & 5 & 5 & 5 \\ \hline
\end{tabular}
}
\label{table.6}
\end{table}

\subsection{Heatmap Comparison}
The heatmap values are from the enhanced text modalities (attended by audio and vision) of MulT and ours. With the the hierarchical architecture of our approach, some words that are not highlighted in MulT are highlighted.

\begin{figure}[!ht]
    \centering
    \includegraphics[scale=0.36]{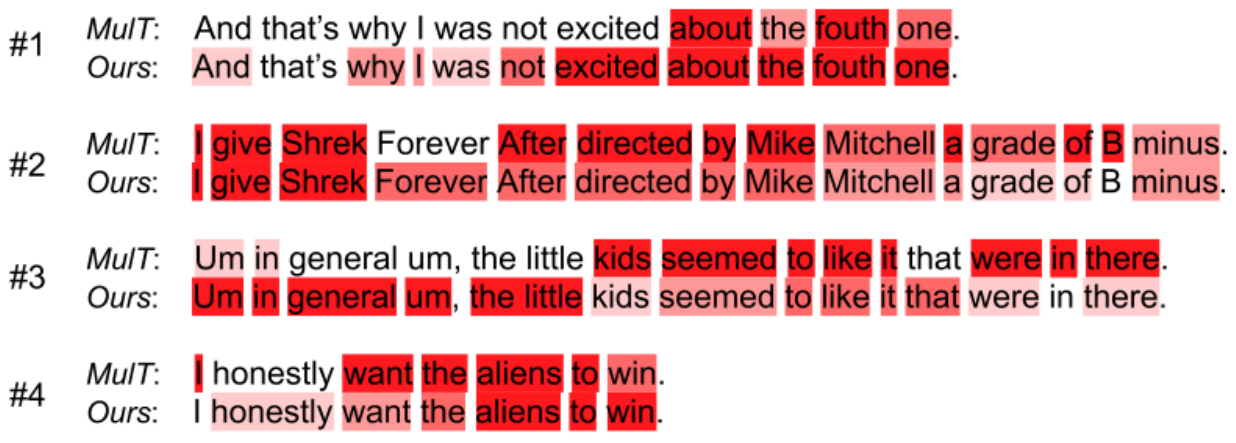}
    \caption{Heatmap comparison of the examples in Table~\ref{example} (Page 7).}
    \label{fig.append}
\end{figure}

\subsection{Primary Modality Selection by DMG with Different Batch Sizes}

Since the training converges with similar patterns to Figure~\ref{fig.urf_prob}, with $T$ dominating followed by $A$ and $V$, we only present the weight variation in the first epoch.

\begin{figure}[!ht]
    \centering
    \includegraphics[scale=0.21]{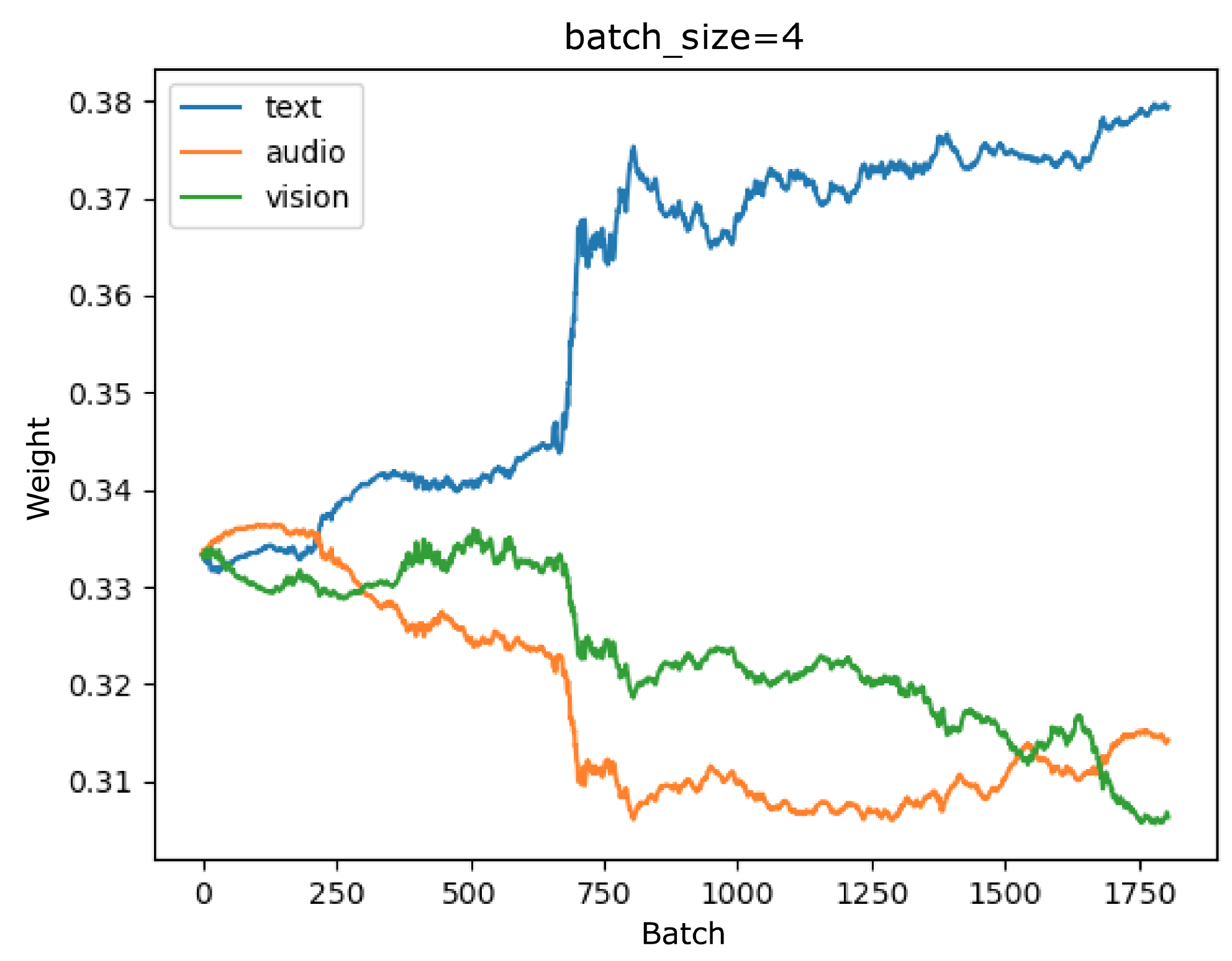}
    \caption{Weight variation in the first epoch on UR-FUNNY using batch size of 4.}
    \label{fig.urf_bs4}
\end{figure}

\begin{figure}[!ht]
    \centering
    \includegraphics[scale=0.21]{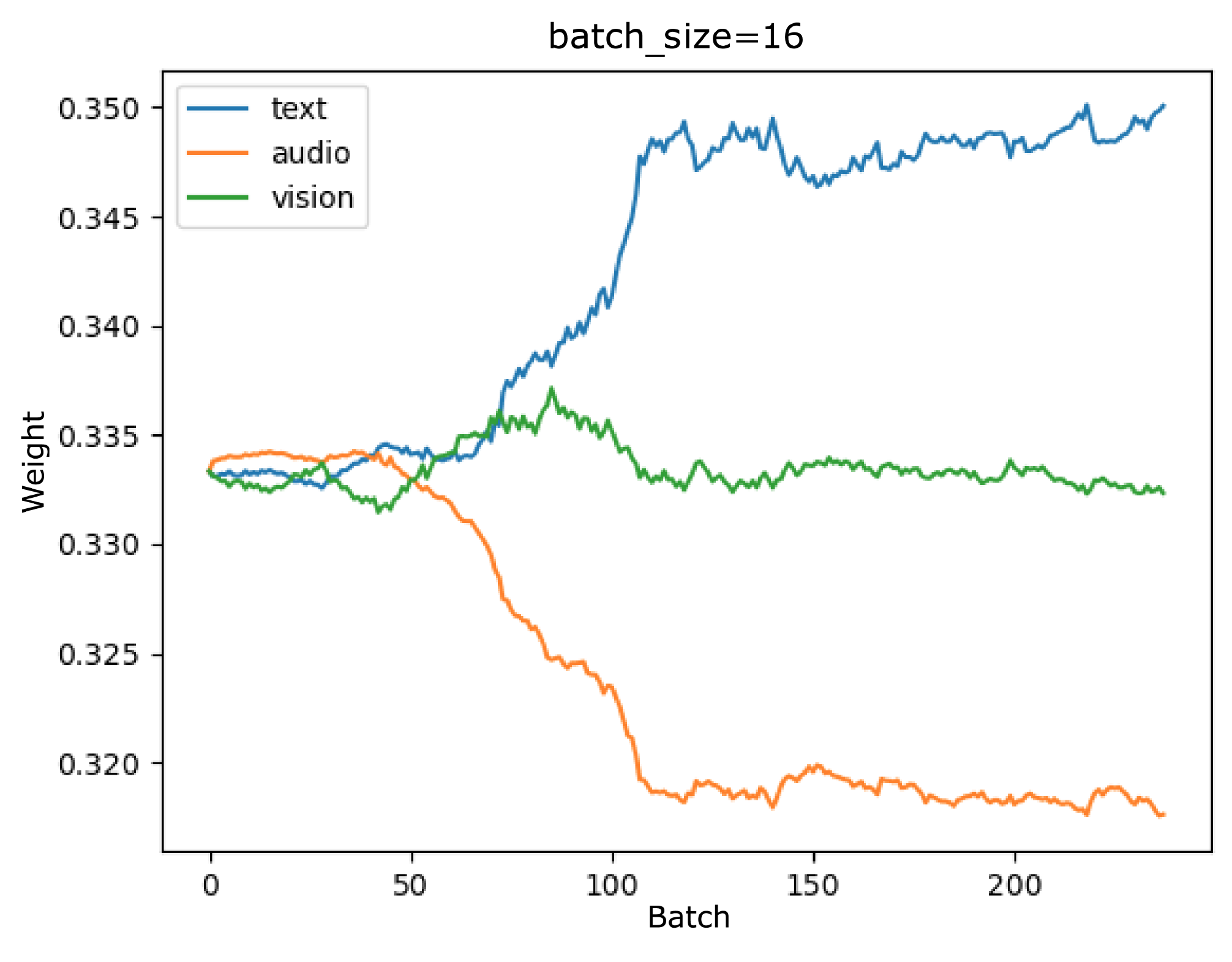}
    \caption{Weight variation in the first epoch on UR-FUNNY using batch size of 16.}
    \label{fig.urf_bs16}
\end{figure}

\begin{figure}[!t]
    \centering
    \includegraphics[scale=0.21]{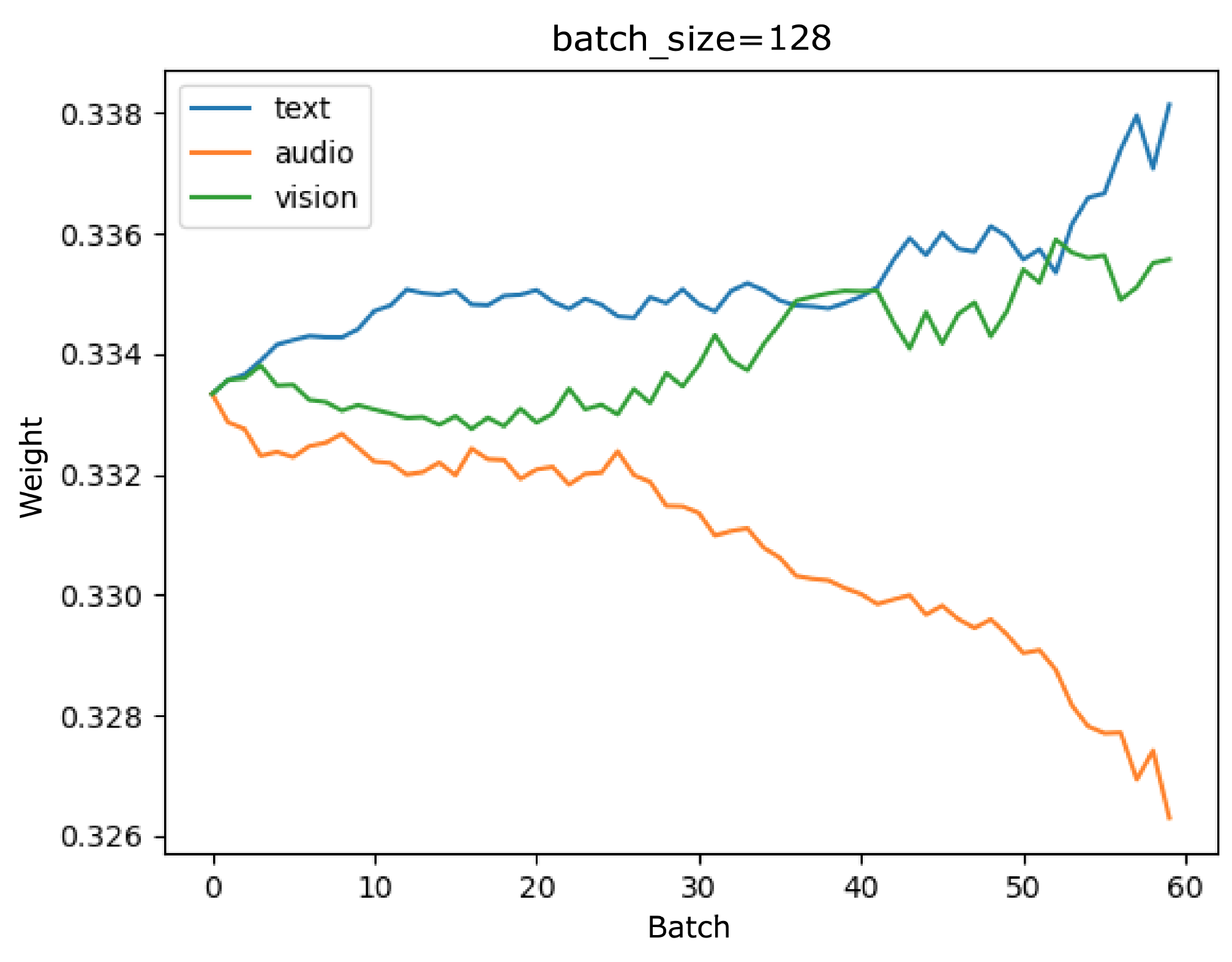}
    \caption{Weight variation in the first epoch on UR-FUNNY using batch size of 128.}
    \label{fig.urf_bs128}
\end{figure}

\end{document}